\documentclass[10pt,twocolumn,letterpaper]{article}

\usepackage{cvpr}              

\usepackage{times}
\usepackage{epsfig}
\usepackage{graphicx}
\usepackage{amsmath}
\usepackage{amssymb}
\usepackage{multirow}
\usepackage{booktabs}
\usepackage{array}
\usepackage{cuted}
\usepackage{capt-of}
\usepackage{xcolor}
\usepackage[accsupp]{axessibility} 

\def\ie{\emph{i.e.}}
\def\eg{\emph{e.g.}}

\def\etc{{\em etc.}}
\newcolumntype{C}[1]{>{\centering\arraybackslash}p{#1}}

\usepackage[pagebackref,breaklinks,colorlinks]{hyperref}

\usepackage[capitalize]{cleveref}
\crefname{section}{Sec.}{Secs.}
\Crefname{section}{Section}{Sections}
\Crefname{table}{Table}{Tables}
\crefname{table}{Tab.}{Tabs.}

\def\cvprPaperID{1478} 
\def\confName{CVPR}
\def\confYear{2022}

\newcommand{\new}[1]{{\color{black}#1}}
\newcommand{\phil}[1]{{\color[rgb]{0.7,0.2,0.2}#1}}

\newcommand{\supp}[1]{{\em #1\/}}

\newcommand{\para}[1]{\vspace{.05in}\noindent\textbf{#1}}

\def\ie{\emph{i.e.}}
\def\eg{\emph{e.g.}}

\def\etc{{\em etc.}}

\begin{document}


\title{Neural Template: Topology-aware Reconstruction and Disentangled Generation of 3D Meshes\vspace{-2mm}}


\author{Ka-Hei Hui$^{1}$\thanks{Joint first authors} \quad Ruihui Li$^{2,1 *}$ \quad Jingyu Hu$^{1}$ \quad Chi-Wing Fu$^{1}$\\
	\hspace{-5mm}$^1$The Chinese University of Hong Kong \hspace{20mm} $^2$ Hunan University\\
	\hspace{10mm}{\tt\small \{khhui,jyhu,cwfu\}@cse.cuhk.edu.hk}\hspace{25mm}{\tt\small liruihui@hnu.edu.cn}\qquad
\vspace{-2mm}
}
%
\maketitle


\begin{abstract}
This paper introduces a novel framework called DT-Net for 3D mesh reconstruction and generation via Disentangled Topology.
Beyond previous works, we learn a topology-aware neural template
specific to each input then deform the template to reconstruct a detailed mesh while
preserving the learned topology.
One key insight is to decouple the complex mesh reconstruction into two sub-tasks: topology formulation and shape deformation.
Thanks to the decoupling,
DT-Net implicitly learns a disentangled representation for the topology and shape in the latent space.
%
Hence, it can enable novel disentangled controls for supporting various shape generation applications,~\eg, remix the topologies of 3D objects, that are not achievable by previous reconstruction works.
%
Extensive experimental results demonstrate that our method\footnote{Code available at~\url{https://github.com/edward1997104/Neural-Template}.} is able to produce high-quality meshes, particularly with diverse topologies, as compared with the state-of-the-art methods.
\end{abstract}
\vspace{-4mm}
\section{Introduction}
\label{sec:intro}


%
%

Polygonal meshes,
as a compact 3D shape representation,
are widely used in many applications, such as modeling, rendering, and animation.
In recent years, generative modeling and reconstruction of 3D meshes has received increasing interest and we may also guide the generative process by using various forms of input,~\eg, images~\cite{wang2018pixel2mesh,pan19deepTopo,groueix2018papier} and point sets~\cite{hao2020dualsdf,deng2020cvxnet,chen2020bsp}.
%
Yet, typical challenges remain---how to deal with the diverse topologies of 3D meshes, and also how to effectively provide high-level controls for new shape generation,~\eg, in a topology-aware manner.


%
To directly reconstruct a 3D mesh,
one popular scheme is to learn to
deform the vertices of an initial template~\cite{ben2018multi,wang2018pixel2mesh,maron2017convolutional,uy2021joint,smith2019geometrics,pan19deepTopo,tang2019skeleton},~\eg, a manually-defined skeleton or a universal sphere, into the target mesh.
However, the topologies of the final reconstructed meshes are typically limited by the template model.
To address this, other works learn to cover a 3D mesh with planar or curved patches~\cite{groueix2018papier,wang2018adaptive};
yet, the visual quality is often tampered due to the patch misalignment, so the resulting meshes often have rough surface appearance.
While other 3D representations such as voxels~\cite{yan2016perspective,girdhar2016learning,wu2017marrnet,tulsiani2017multi,wu2018learning,yang2018learning}, point clouds~\cite{fan2017point,jiang2018gal,achlioptas2018learning}, and implicit functions~\cite{park2019deepsdf,michalkiewicz2019implicit,atzmon2020sal,gropp2020implicit,takikawa2021neural} have been explored, these representations typically require conversions to meshes via a post-processing step for supporting visual applications.

\begin{figure}[t]
\centering
\includegraphics[width=0.97\linewidth]{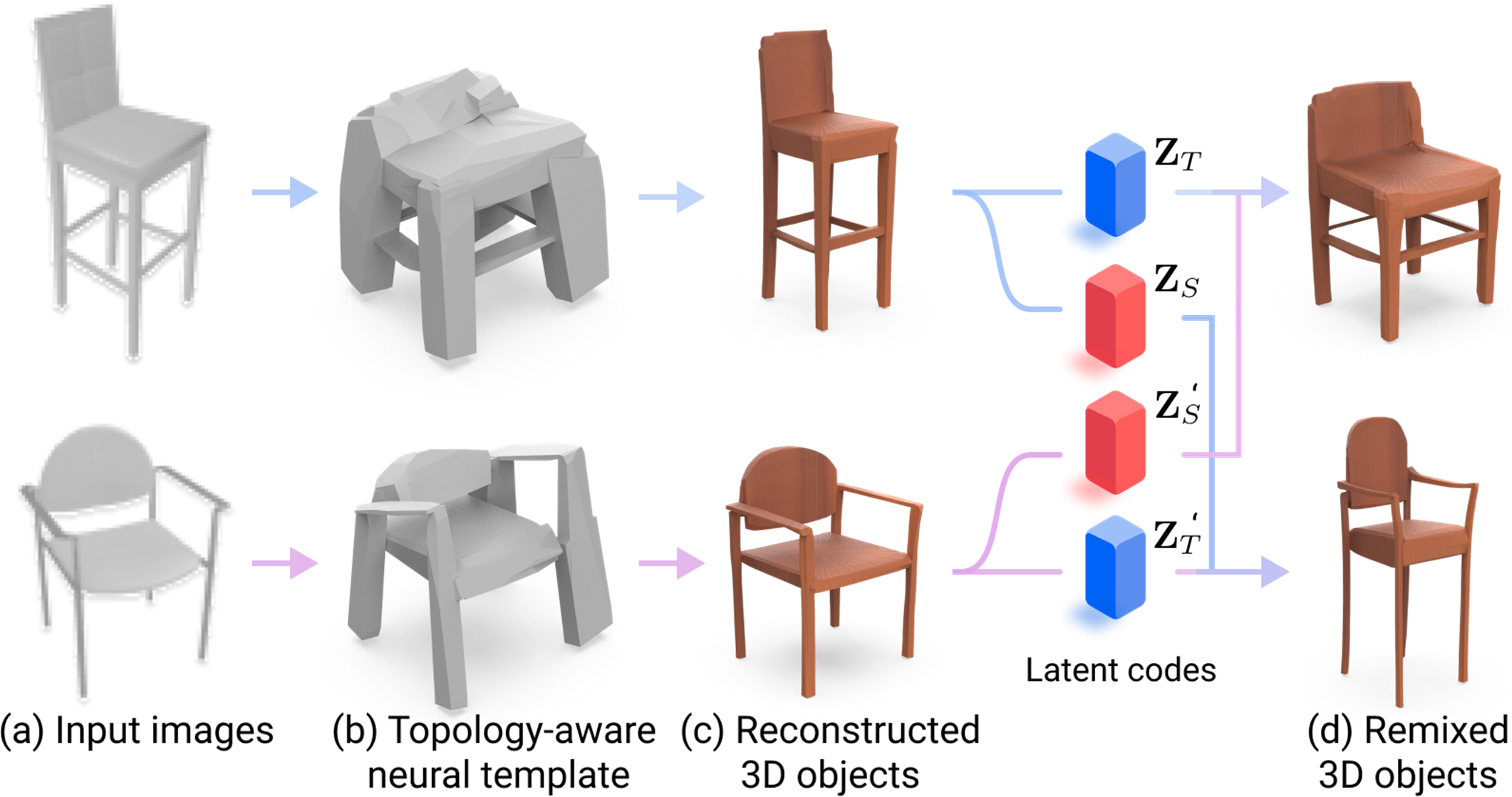}
\vspace{-1.7mm}
\caption{Our DT-Net learns to construct a {\em topology-aware neural template\/} (b) adapted to the input (a) and then deform it towards an accurate 3D mesh while preserving the initial (learned) topology.
This decoupled design enables a disentangled latent representation of {\em \textcolor[rgb]{0.00,0.07,1.00}{topology}\/} ($\mathbf{Z}_T$) and {\em \textcolor[rgb]{1.00,0.07,0.00}{shape}\/} ($\mathbf{Z}_S$), promoting controllable 3D mesh generation,~\eg, remixing codes for object re-synthesis.
}
\label{fig:teaser}
\vspace{-3mm}
\end{figure}

Another drawback is that most works focus on capturing the mesh geometry
directly in a single step, without providing high-level interpretability,~\eg, structure or topology.
So, it is particularly hard
to control the mesh generation process.
%
Some recent works tried to address this shortcoming by generating objects using parts and parts composition,~\eg,
in terms of voxels~\cite{wu2019sagnet}, point clouds~\cite{mo2019structurenet,wu2020pq}, and meshes~\cite{gao2019sdm}.
While the approach allows certain part-aware generation, these works highly rely on the availability and the quality of the extra parts annotations.

In this paper, we present a novel framework, namely \emph{DT-Net}, for 3D mesh reconstruction and generation via disentangled topology (DT).
Distinctively, DT-Net enables the reconstruction of high-quality 3D meshes with diverse topologies,
well-adapting to the input,~\eg, images or voxels.
Also, our novel design facilitates controllability
in the generative process, since DT-Net implicitly learns a disentangled latent representation for the topology and shape.
Therefore, we can achieve disentangled mesh generations with separate topology and shape manipulations.

Figure~\ref{fig:teaser} illustrates the pipeline of DT-Net.
Beyond previous works, we learn
a topology-aware neural template (\eg, genus of chairs) that fits each input then deform the template to reconstruct a detailed mesh.
%
A key insight behind our design is that we decouple the mesh reconstruction into two sub-tasks:
(i) topology formation for adapting various topologies;
and (ii) shape deformation for reconstructing accurate objects while respecting their initial topologies.
Our decoupling scheme eases the learning process and accounts for the topology, while enhancing the reconstruction quality and enriching mesh generation with diverse topologies.
Another important design is that we extract a topology code (blue) and a shape code (red) from the input, to guide the learning of the two decoupled sub-tasks, respectively.
%
%
By doing so,
two key aspects of 3D objects, topology and shape, can be jointly learned to ensure the reconstruction plausibility, while being disentangled in the latent space, for enabling novel disentangled controls in the mesh generation process; see Figure~\ref{fig:teaser} (right).
Please refer to Section~\ref{subsec:framework} for further elaborations on our framework.

Method-wise, we design an end-to-end framework with
%
the topology-learning module to first learn to produce a topology-aware neural template composed of convexes.
%
To decouple topology learning and shape learning, we learn a family
of invertible maps~\cite{yang2019pointflow,Gupta2020neural} to maintain the topology between the neural template and the final reconstructed object.
%
Also, we propose to use a dual (implicit and explicit) representation for the neural template, so
it can be trainable via the implicit functions and extractable as polygonal meshes at the inference.
%
%
Importantly, our approach can directly learn the topology-aware neural template without intermediate topology annotations, while well-aligning it with an inversely-deformed version of the ground-truth mesh.

Both quantitative and qualitative results show that DT-Net enables the reconstruction of high-quality meshes with diverse topologies, performing favorably over the state of the arts.
%
Further, our method supports various generative applications via disentangled controls, which cannot be achieved by existing reconstruction-based methods.

\section{Related Work}
\label{sec:rw}

Learning-based shape synthesis and analysis have attracted increasing research interest recently,
benefiting from the availability of large shape collections~\cite{chang2015shapenet,mo2019partnet} and advances in the design of generative neural networks.
In this section, we briefly review the recent advances in 3D reconstruction and generative modeling.
We first focus on the mesh representation of object surfaces, learned explicitly or implicitly, and then discuss related works on shape abstraction and disentangled representation learning.

\para{Explicit surface representation}
has been extensively studied for 3D voxels~\cite{yan2016perspective,girdhar2016learning,zhu2017rethinking,wu2017marrnet,tulsiani2017multi,wu2018learning,yang2018learning}, octrees~\cite{riegler2017octnet,hane2017hierarchical,tatarchenko2017octree,wang2018adaptive}, and point clouds~\cite{fan2017point,jiang2018gal,achlioptas2018learning,li2021sp}.
%
However, these representations are usually restricted to low resolutions and lack an explicit topology for detailed shape reconstruction.
%

In contrast, polygonal mesh is an efficient and continuous surface representation with local topological information explicitly defined by the connections between vertices.
Since the learning of connection relations is challenging, most mesh-based approaches strive to learn a vertex-based deformation of an initial mesh template with graph convolutions~\cite{wang2018pixel2mesh}, MLPs~\cite{tulsiani2020implicit,groueix2018papier}, or neural ODEs~\cite{Gupta2020neural}.
These initial meshes are either searched from
a set of CAD models~\cite{huang2015single,su2014estimating, kong2017using,pontes2018image2mesh}, customized category-based templates~\cite{kolotouros2019convolutional,zuffi2018lions}, or category-agnostic meshes~\cite{wang2018pixel2mesh,smith2019geometrics,pan19deepTopo,tang2019skeleton,groueix2018papier,shi2021geometric}, such as a genus-zero ellipsoid or 2D planer patches.

While these mesh-based methods achieve finer reconstructions,
the topologies of the generated objects are constrained by the template models that they deform from.
Instead of manually or explicitly defining a template, we learn to produce a topology-aware neural template adapted to the input, promoting a high-quality reconstruction with varying topologies.
Particularly, the disentangled topology also enables our method to support controllable shape generation, which is not achievable by the existing methods.

\begin{figure*}[t]
	\centering
	\includegraphics[width=0.99\linewidth]{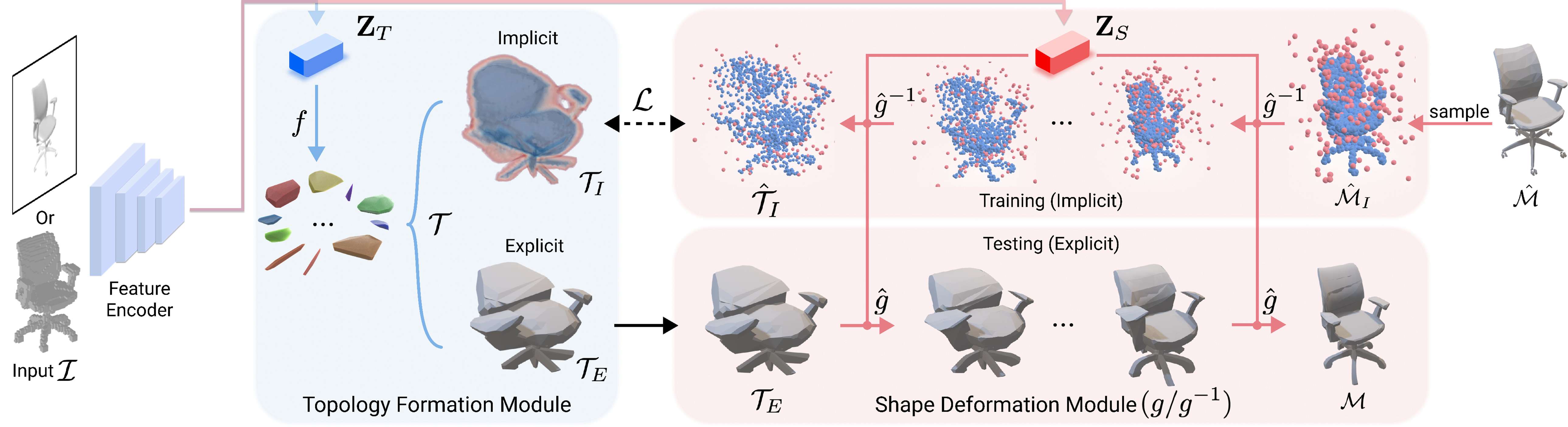}
	\vspace{-1.5mm}
	\caption{Overview of our DT-Net framework.
		Given an input, either a single-view image or 3D voxels, the encoder predicts two separate feature vectors: topology code $\mathbf{Z}_{T}$ and shape code $\mathbf{Z}_{S}$.
		Then, from $\mathbf{Z}_{T}$, we produce neural template $\mathcal{T}$ with an implicit representation $\mathcal{T}_{I}$ and an explicit representation $\mathcal{T}_{E}$ through $f$ in the topology formation module.
		During the inference, we progressively deform $\mathcal{T}_{E}$ by function $g$ in the shape deformation module conditioned on $\mathbf{Z}_{S}$ to obtain the final reconstructed shape $\mathcal{M}$.
	    	We supervise the training by using the occupancy pairs $\mathcal{\hat{M}}_{I}$ sampled in the shape space from the ground-truth mesh $\mathcal{\hat{M}}$. Also, we inversely map $\mathcal{\hat{M}}_{I}$ to the topology space by inverse function $g^{-1}$ to produce $\mathcal{\hat{T}}_{I}$ for evaluating the corresponding occupancy on the implicit template representation $\mathcal{T}_{I}$,
	    	promoting a correct alignment between the learned topology $\mathcal{T}_{I}$ and the inversely-deformed shape $\mathcal{\hat{T}}_{I}$ by the loss $\mathcal{L}$.}
	\label{fig:framework}
	\vspace{-3.5mm}
\end{figure*}

\para{Implicit surface representation} models a 3D shape as a level set
of discrete volume or continuous field, from which we can extract a surface mesh,~\eg, via iso-surfacing~\cite{lorensen1987marching}.
%
From an input image, these methods extract a context vector then train a neural network to predict a signed distance field~\cite{park2019deepsdf,michalkiewicz2019implicit,atzmon2020sal,gropp2020implicit,takikawa2021neural}
or occupancy probabilities~\cite{mescheder2019occupancy,chen2019learning} for 3D reconstruction.
Some recent works attempted to adopt extra information, e.g., camera pose~\cite{xu2019disn,xu2020ladybird,li2021d2im} and shape skeletons~\cite{tang2019skeleton,tang2021skeletonnet}, to enhance the 3D reconstructions.

While these methods improve the reconstruction quality, they lack
interpretability on the 3D structure or topology.
In this work, we propose to implicitly learn a disentangled representation for the topology and shape, facilitating novel controls on the 3D mesh generation process.
%
%

\para{Shape abstraction} aims to coarsely approximate shapes with
few primitives like
cuboids~\cite{tulsiani2017learning,niu2018im2struct,zou20173d,sun2019learning}, 
super-quadrics~\cite{paschalidou2019superquadrics,remelli2020meshsdf,riegler2017octnetfusion}, and spheres~\cite{hao2020dualsdf}.
%
Recent works~\cite{genova19implicit,genova2020local,paschalidou2021neural,deng2020cvxnet,chen2020bsp} also
leverage a structured set of implicit primitives to compose shapes.
With primitives defined explicitly, these methods enable a direct extraction of 3D meshes.
%
We draw inspiration from them to design our framework.

\para{Disentangled representations} have been widely studied in image
generation, allowing manipulations separately in different aspects,~\eg, texture style~\cite{lee2018diverse,huang2018multimodal}, facial
attribute~\cite{chen2016infogan,karras2019style},~\etc~
For disentanglement in 3D shapes~\cite{abrevaya2019decoupled,zhou2020unsupervised,aumentado2019geometric}, some existing works focus on specific categories such as human faces and animal bodies.
Alternatively, with additional part annotations~\cite{mo2019partnet}, some recent works~\cite{wu2019sagnet,mo2019structurenet,wu2020pq,gao2019sdm,jie20dsgnet} tried to
achieve certain part-based disentanglement by encoding parts separately and composing parts into objects via decoding.
Yet, the above works depend greatly on the availability and quality of the parts or structures annotations.
In contrast, our new approach decouples the reconstruction process into topology formation and shape deformation, promoting topology and shape disentanglement automatically, \emph{without} requiring these annotations as supervision.

\para{Other related works.} Our work shares some conceptual similarities with neural cages~\cite{yifan2020neural}, as both predict an input-adaptive mesh (template (our) or cage~\cite{yifan2020neural}) for further deformation.
Yet, our objectives and applications are very different.
Also, we noticed few recent works~\cite{zheng2021deep,deng2021deformed} learn a shared implicit-field-based template per category for modeling dense correspondence among shapes. Different from them, we learn a topology template adapted to each input for enhancing 3D reconstructions with diverse topologies.

\section{Method}
\label{sec:method}

\subsection{The DT-Net Framework}  
\label{subsec:overview}


Figure~\ref{fig:framework} shows an overview of our DT-Net framework, which consists of two modules, the {\em topology formulation module\/} and {\em shape deformation module\/}.
Given input $\mathcal{I}$, which can be a 2D image or a 3D voxelized data, DT-Net first encodes it to produce two separate feature vectors, the {\em topology code\/} $\mathbf{Z}_{T}$ and {\em shape code\/} $\mathbf{Z}_{S}$.
To match the given input, the topology formulation module takes $\mathbf{Z}_{T}$ to generate topology-aware neural template $\mathcal{T}$, whereas the shape deformation module takes $\mathbf{Z}_{S}$ as a guidance to refine $\mathcal{T}$ to produce the final output $\mathcal{M}$ with geometric details.
In the topology formation module, we learn function $f$ to compose $\mathcal{T}=f(\mathbf{Z}_{T})$ using a set of learned convexes.
Then in the shape deformation module, we learn an {\em invertible\/} homeomorphic flow function $g$ to progressively deform $\mathcal{T}$ towards $\mathcal{M}=g(\mathbf{Z}_{S},\mathcal{T})$.
Note that both $f$ and $g$ are implemented as neural networks; see the details in Section~\ref{subsec:network}.

Very importantly, we design $\mathcal{T}$ to have a {\em dual representation\/}; see again Figure~\ref{fig:framework}.
The {\em explicit representation\/} $\mathcal{T}_{E}$  is in the form of 3D meshes (\ie,~vertices and faces on object surface), while the {\em implicit representation\/} $\mathcal{T}_{I}$ is in the form of an implicit field (\ie, an occupancy function that indicates whether any given query point is inside/outside the object).
By this design, the training of DT-Net (essentially $f$ and $g$) can be differentiable via the implicit representations (see the top branch in Figure~\ref{fig:framework});
see more details later in this section.
On the other hand, during the inference (see the bottom branch), $\mathcal{T}_{E}$ and $\mathcal{M}$ can be directly extracted as explicit meshes using the trained $f$ and $g$.

Further, we refer to the 3D space of $\mathcal{T}$ as the \emph{topology space} and the 3D space of $\mathcal{M}$ as the \emph{shape space}.
To obtain a continuous gradient between the two spaces, we learn inverse function $g^{-1}$ from the shape space to topology space,~\ie, $\hat{\mathcal{T}_{I}} = g^{-1}(\mathbf{Z}_{S},\hat{\mathcal{M}_{I}})$.
As shown in Figure~\ref{fig:framework} (top right), during the training, we sample occupancy field $\mathcal{\hat{M}}_{I}$ (\ie,~point coordinates and occupancy values) from the ground-truth mesh $\hat{\mathcal{M}}$ in the shape space.
%
%
Using $\mathcal{\hat{M}}_{I}$, we can then construct $\hat{\mathcal{T}_{I}}$ using $g^{-1}$ and formulate a regularization in the topology space as $\mathcal{L}(\hat{\mathcal{T}_{I}},\mathcal{T}_{I})$:

\vspace*{-3mm}
\begin{equation}
\label{eq:topology}
\min_{f,g}~\mathcal{L}(g^{-1}(\mathbf{Z}_{S},\hat{\mathcal{M}_{I}}),f(\mathbf{Z}_{T})) .
\end{equation}

\vspace*{-1mm}
\noindent
This optimization function defines how well the inversely-transformed implicit shape $\hat{\mathcal{T}_{I}}$ (from $g^{-1}$) aligns with the composed implicit neural template $\mathcal{T}_{I}$ (from $f$).
%
%
%
%
%
%

\para{Implicit representation} $\mathcal{T}_{I}$ can be derived by the bijective mapping $g$ between the topology and shape spaces. Since $g: R^3 \leftrightarrow R^3$ is a point-wise continuous function,
$\mathcal{T}$, or more specifically $\mathcal{T}_{I}$, can be derived by using

\vspace{-3mm}
\begin{equation}
\label{eq:implicit}
	O(\mathcal{T}_{I}, \hat{\mathcal{T}_{I}}) =
	O(\mathcal{T}_{I}, g^{-1}(\mathbf{Z}_{S}, p))
\end{equation}

\noindent
where $\{p\}$ are sample points in $\mathcal{\hat{M}}_{I}$.
To evaluate point $p$ relative to $\mathcal{\hat{M}}$, it suffices to find whether the transformed point $g^{-1}(\mathbf{Z}_{S}, p)$ is inside or outside the surface of $\mathcal{T}$, calculated via an occupancy function $O(\cdot)$.
In other words, points that are originally inside (outside) $\mathcal{\hat{M}}$, after an inverse transformation, should also be inside (outside) the $\mathcal{T}$ as well.

\para{Explicit representations} $\mathcal{T}_{E}$ and $\mathcal{M}$ are 3D meshes that have the same face set $F$ but different vertex sets $V_{\mathcal{T}}$ and $V_{\mathcal{M}}=g(\mathbf{Z}_{S},V_{\mathcal{T}})$, respectively.
Here, functions $g$ and $g^{-1}$ map between corresponding vertices in $V_{\mathcal{T}}$ and $V_{\mathcal{M}}$.
%
So, the final reconstructed object $\mathcal{M}$ can be obtained through a deformation (function $g$) from the template mesh $\mathcal{T}_{E}$.
We can extract $\mathcal{T}_{E}$ by grouping a set of learned primitives, such that we can flexibly represent 3D objects of various topologies.
More details will be given in Section~\ref{subsec:network}.

\subsection{Framework Design}
\label{subsec:framework}

Before elaborating on the details of the DT-Net framework, we first discuss the key ideas in framework design.
\begin{itemize}
\vspace*{-2.3mm}
\item[(i)]
\emph{Topology-aware learning}.
%
We learn to produce neural template $\mathcal{T}$ whose topology specifically follows the input $\mathcal{I}$, instead of manually defining a template as previous works.
To adapt $\mathcal{T}$ for varying topologies, we produce it
by composing geometric primitives, which are based on well-defined implicit and explicit representations~\cite{chen2020bsp,deng2020cvxnet}.
Thus, $\mathcal{T}$ is {\em trainable\/} via implicit functions and {\em extractable\/} directly as explicit meshes.
%
%
\vspace*{-2.3mm}
\item[(ii)]
\emph{Topology-preserved deformation}.
To decouple topology learning and shape learning, we preserve the topology of the neural template $\mathcal{T}$ while deforming it to form the output mesh.
Particularly, we learn a family of invertible maps~\cite{yang2019pointflow,Gupta2020neural} between the topology space and shape space, such that we can impose various
constraints on $\mathcal{T}$ from $\hat{\mathcal{M}}$ for efficiently computing its implicit and explicit representations.
%
\vspace*{-2.3mm}
\item[(iii)]
\emph{Without topology annotations.}
DT-Net learns to produce the topology-aware neural template $\mathcal{T}$ directly from input $\mathcal{I}$ and ground-truth mesh $\mathcal{\hat{M}}$ {\em without\/} requiring topology annotations as the intermediate supervision.
%
%
We achieve so by inversely mapping $\mathcal{\hat{M}}$ from the shape space into the topology space,~\ie, by inversely deforming sample points $\mathcal{\hat{M}}_{I}$ into $\mathcal{\hat{T}}_{I}$.
%
So, DT-Net can learn to produce $\mathcal{T}_{I}$ in an unsupervised manner by precisely aligning $\mathcal{\hat{T}}_{I}$ with the learned $\mathcal{T}$, as Eq.~\eqref{eq:topology}.
%
\vspace*{-2.3mm}
\item[(iv)]
\emph{Topology \& shape disentanglement}.
Also, we provide controllability in the generation by injecting the topology code $\mathbf{Z}_{T}$ and shape code $\mathbf{Z}_{S}$ into the training of $f$ and $g$, respectively.
By this means, topology and shape are jointly learned to ensure plausible reconstructions, while being as disentangled as possible in the latent space.
This design provides a family of novel high-level controls, such as manipulating a mesh by modifying its shape code while preserving its topology code;
examples will be presented in Section~\ref{subsec:generation}.
%
\end{itemize}

\subsection{Network Architecture}
\label{subsec:network}


\para{Topology formation module.}
We learn function $f$ to map topology code $\mathbf{Z}_{T}$ to neural template $\mathcal{T}=f(\mathbf{Z}_{T})$.
Inspired by~\cite{tulsiani2017learning,paschalidou2019superquadrics,paschalidou2021neural,deng2020cvxnet,chen2020bsp}, 
we propose to compose the topology-aware neural template via a union of geometric primitives.
As referred to the key idea (i) in Section~\ref{subsec:framework}, we adopt the formulation in~\cite{chen2020bsp} to group a collection of convex polyhedra to assemble the implicit field of the neural template.

Specifically, given $\mathbf{Z}_{T}$, we implement $f$ using multi-layer perceptrons that first predict the parameters to define the various hyperplanes $\mathcal{H} \in R^{N_{h}*4}$ (\ie, $ax+by+cz+d=0$) then group these planes to form a set of convexes $\mathcal{C}$ via a learnable binary matrix $\mathbf{B} \in R^{N_{h}*N_{c}}$ (a selective mask), where
$N_{h}$ and $N_{c}$ denote the number of hyperplanes and convexes, respectively.
Lastly, these convexes are assembled to form the neural template $\mathcal{T}$.
This formulation enables an explicit representation $\mathcal{T}_{E}$ (\ie, a union of the convexes) and also an implicit representation $\mathcal{T}_{I}$ (\ie, a scalar function $O(\cdot)$ in Eq.~\eqref{eq:implicit} for indicating the occupancy: a given point is inside/outside these convexes).

\begin{figure*}[t]
	\centering
	\includegraphics[width=0.95\linewidth]{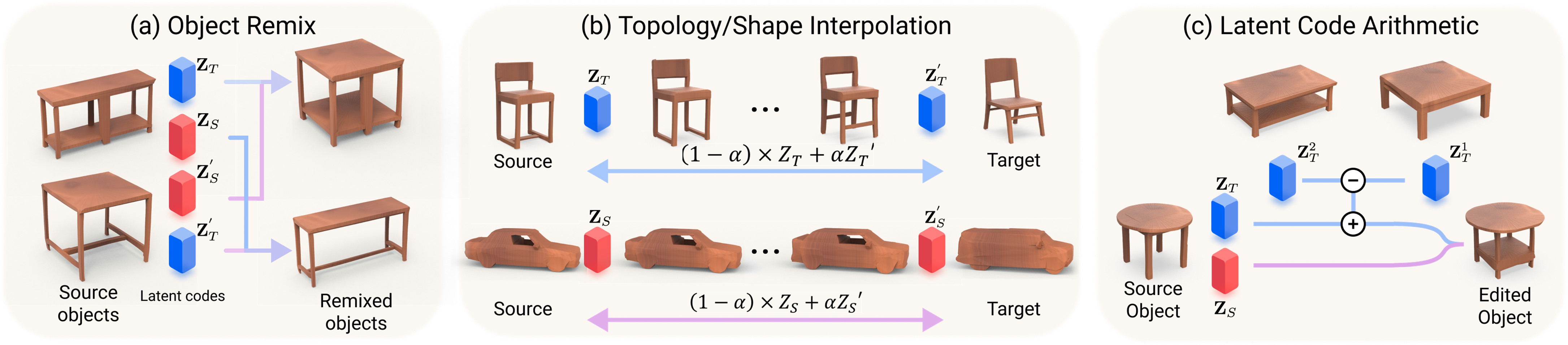}
\vspace{-2.2mm}
	\caption{Our DT-Net framework learns a disentangled representation for the topology and shape, thus facilitating novel generation applications via disentangled manipulation on the topology code $\mathbf{Z}_{T}$ and/or the shape code $\mathbf{Z}_{S}$,~\eg,
		(a) remix the shape and topology of two different objects;
		(b) object interpolation by manipulating the topology/shape code; and
		(c) arithmetic operations in the latent space.}
	\label{fig:manipulation_demo}
	\vspace{-3.1mm}
\end{figure*}

\para{Shape deformation module.}
We learn invertible deformation function $g$ that preserves the topology between output object and learned neural template; see key idea (ii) above.
Given $\mathbf{Z}_{S}$, it learns to progressively deform the neural template $\mathcal{T}$ towards the detailed surface $\mathcal{M}=g(\mathbf{Z}_{S},\mathcal{T})$.
%
%

Specifically, we adopt the neural ordinary differential equation module (NODE) in~\cite{yang2019pointflow,Gupta2020neural} to achieve a continuous deformation on the topologies.
It defines an invertible transformation $g: R^3 \leftrightarrow R^3$ via a parameterized ODE $p_T = g(\mathbf{Z}_{S},p_0) = p_0 + \int_{0}^{T} \hat{g}(\mathbf{Z}_{S},p_t)~dt$, where $p_0$ and $p_T$ are input to and output from neural network $\hat{g}$ (\ie, $[x,y,z]$), and $T$ is a hyperparameter that denotes the number of deformation steps from $p_0$ to $p_T$.
This integration is approximated using numerical solvers, while its gradient can be computed by using the adjoint method proposed in~\cite{chen2018neuralode}.
%
Due to the diffeomorphic nature of $g$, we can then preserve the general topology of $\mathcal{T}$ in the deformation process.

\subsection{Network Training}
\label{sec:train}

Without topology annotations, as mentioned in Section~\ref{subsec:overview}, we propose to train DT-Net via $\mathcal{L}(\hat{\mathcal{T}_{I}},\mathcal{T}_{I})$.
The joint-optimization function is composed of two terms:
\begin{equation}
    \label{eq:loss}
	\mathcal{L}(\hat{\mathcal{T}_{I}},\mathcal{T}_{I}) = \mathcal{L}_{\text{align}} + \mathcal{L}_{\mathbf{B}},
\end{equation}
where $\mathcal{L}_{\text{align}}$ encourages a correct alignment between the
learned
topology $\mathcal{T}_{I} = f(\mathbf{Z}_{T})$
and the inversely-deformed shape $\hat{\mathcal{T}_{I}} = g^{-1}(\mathbf{Z}_{S},\hat{\mathcal{M}_{I}})$.
Also, we adopt the sparsity term $\mathcal{L}_{\mathbf{B}}$ in~\cite{chen2020bsp} to encourage the learned topology to be composed by a sparse set of convexes.


%
Specifically, $\mathcal{\hat{M}}_{I} = \{p'_i, o_i\}^{N_p}_{i=1}$ denotes $N_p$ occupancy pairs sampled from the shape space of ground-truth mesh $\hat{\mathcal{M}}$;
%
$p'_i$ is the $i$-th sample point coordinates, and 
$o_i$$=$$1$ ($o_i$$=$$0$) indicates that 
$p'_i$ is inside (outside) the object.
By inversely mapping $p'_i$ into the topology space as $p_i =  g^{-1}(\mathbf{Z}_{S}, p'_i)$, we then obtain $\mathcal{\hat{T}}_{I} = \{p_i, o_i\}^{N_p}_{i=1}$ as the intermediate signal to optimize topology learning function $f$.
For each query point $p_i \in \mathcal{\hat{T}}_{I}$ and associated ground-truth occupancy value $o_i$,  $\mathcal{L}_{\text{align}}$ measures the difference between $O(\mathcal{T},p_i)$ and $o_i$, promoting the network to predict the right occupancy value.

To ease the gradient flow, we adopt the two-stage training strategy in~\cite{chen2020bsp}:
stage 1 (continuous) computes a relax approximation $\mathcal{L}^{\text{con}}(\hat{\mathcal{T}_{I}},\mathcal{T}_{I})$ from $\mathcal{T}_{I}$ to $\hat{\mathcal{T}_{I}}$, then
stage 2 (discrete) promotes an accurate alignment $\mathcal{L}^{\text{dis}}(\hat{\mathcal{T}_{I}},\mathcal{T}_{I})$ between $\mathcal{T}_{I}$ and $\hat{\mathcal{T}_{I}}$.
Specifically,
$\mathcal{L}^{\text{con}}_{\text{align}}$ adopts a least-squares model to approximate the ground-truth occupancy value $o_i$, whereas $\mathcal{L}^{\text{dis}}_{\text{align}}$ adopts binary cross entropy to encourage the output occupancy value to be discrete as $o_i$:
%
\vspace*{-1.5mm}
\begin{eqnarray}
  \label{eq:rec1}
	\mathcal{L}^{\text{con}}_{\text{align}} &=&  \frac{1}{N_p}\sum_{i=1}^{N_p}{(O(\mathcal{T}_{I},p_i) - o_i)^2} \nonumber \\
	\text{and} \ \ \ \mathcal{L}^{\text{dis}}_{\text{align}} &=&   \frac{1}{N_p} \sum_{i=1}^{N_p} \big[ o_i* \max(O(\mathcal{T}_{I},p_i), 0)  \nonumber \\ &&+  (1 - o_i) * (1 - \min(O(\mathcal{T}_{I},p_i), 1)) \big]. \nonumber
\end{eqnarray}

\subsection{Shape Generation with Controllability}
\label{subsec:generation}

With the disentangled representation,~\ie, the topology code $\mathbf{Z}_{T}$ and shape code $\mathbf{Z}_{S}$, DT-Net enables novel forms of 3D object manipulations,
%
opening up new possibilities for high-level object generation and re-synthesis:

%

\begin{figure}[t]
	\centering
	\includegraphics[width=0.9\columnwidth]{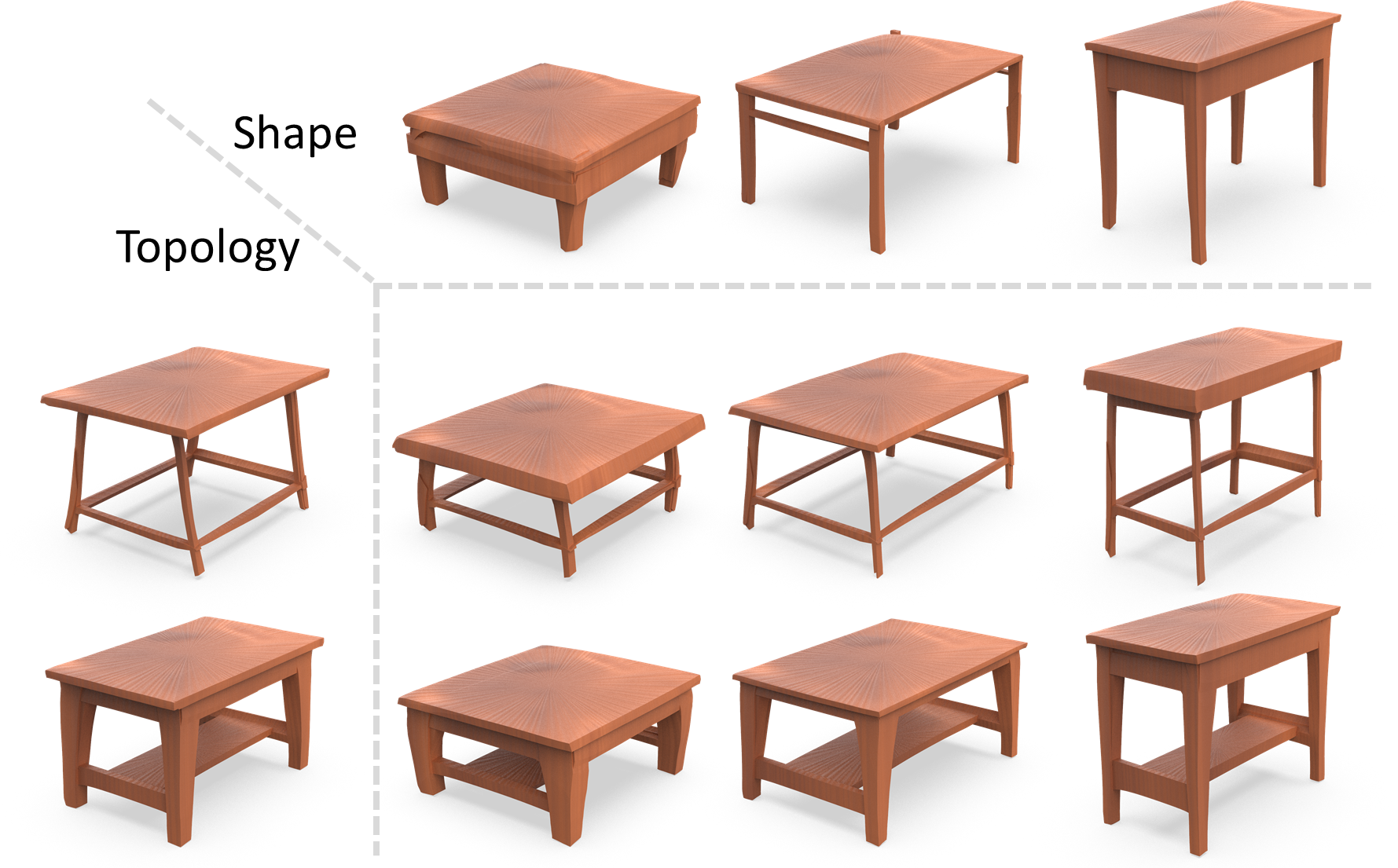}
\vspace{-2mm}
	\caption{Remixing shape and topology.
	Objects on top provide the shape codes and objects on the left provide the topology codes.}
	\label{fig:remix}
	\vspace{-2mm}
\end{figure}

\begin{figure}[t]
	\centering
	\includegraphics[width=0.95\linewidth]{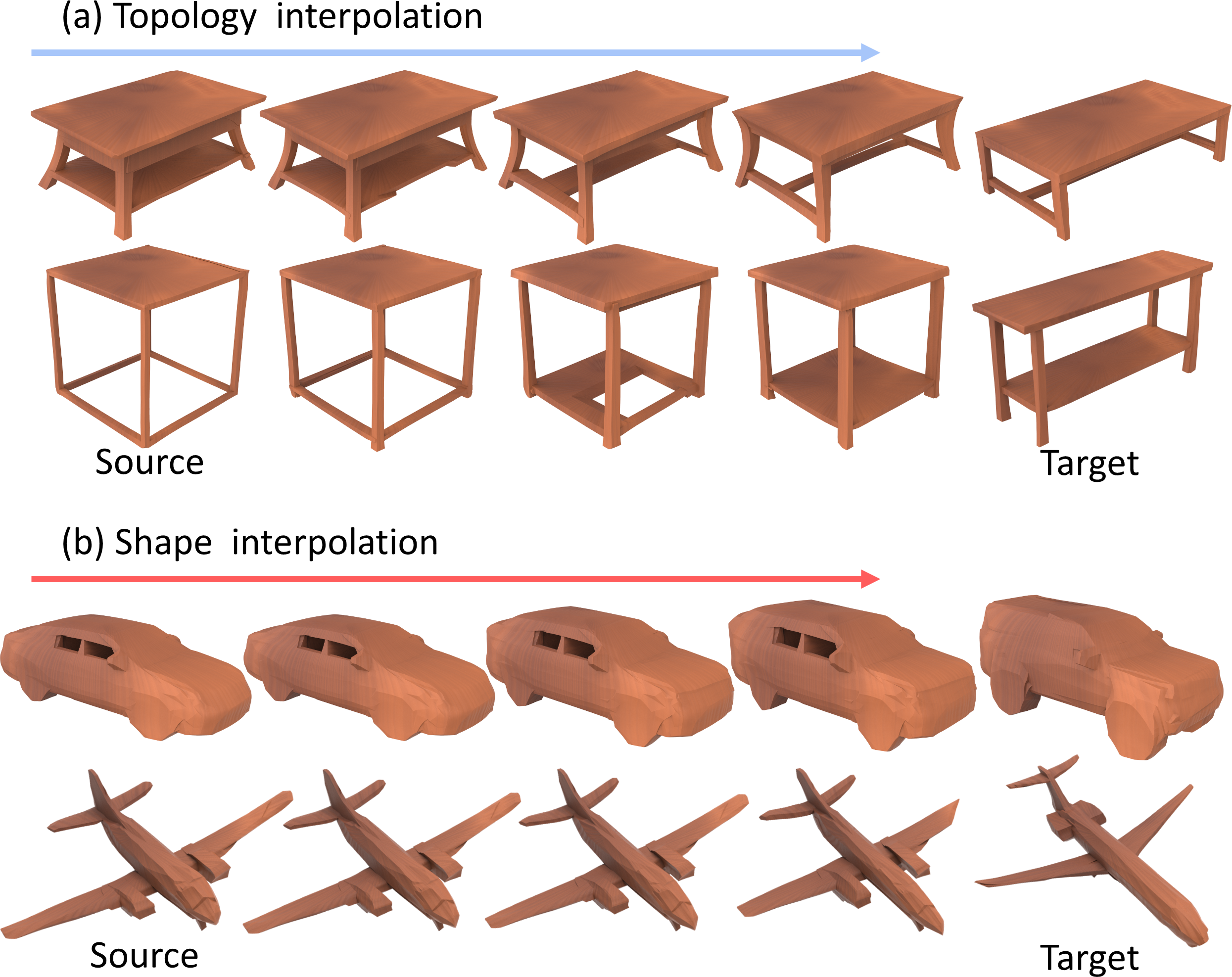}
\vspace{-2mm}
	\caption{Object interpolation separately on topology (top) and shape (bottom).
	Note the smooth transitions achieved by DT-Net.}
	\label{fig:interpolation}
	\vspace{-3.5mm}
\end{figure}

\begin{figure*}[t]
	\centering
	\includegraphics[width=0.95\linewidth]{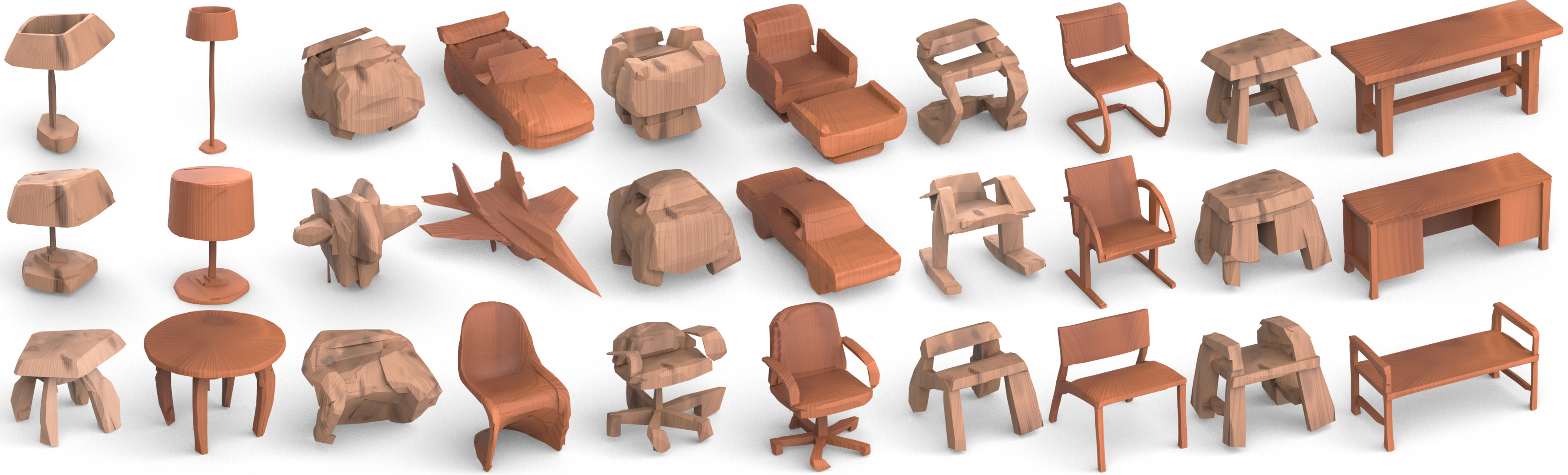}
\vspace{-3mm}
	\caption{Galleries showcasing the results produced by our DT-Net. Each pair shows the learned topology-aware neural template (left) and the associated reconstructed object (right).
The produced objects cover various shapes and diverse topologies, ranging from smooth surface (\eg,~car and lamp), to complex geometry (\eg,~chair and airplane). It is observed that the neural template visually appears like a coarse version of the final shape, even without regularizing the amplitude of the deformation module.}
	\label{fig:gallery}
	\vspace{-3mm}
\end{figure*}

\begin{figure}[!t]
	\centering
	\includegraphics[width=0.925\linewidth]{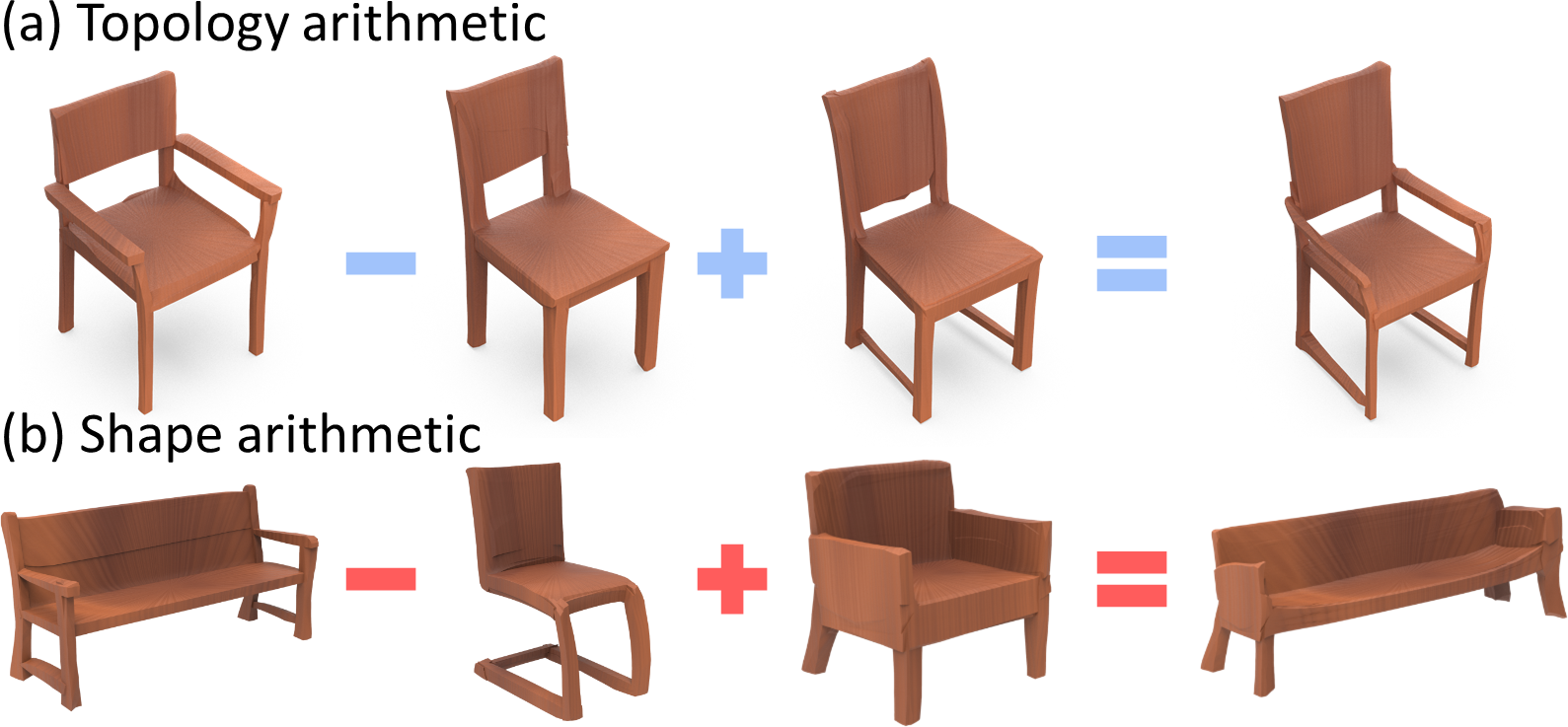}
\vspace{-3mm}
	\caption{Arithmetic operations between different objects.}
	\label{fig:arithmetic}
	\vspace{-3.5mm}
\end{figure}

\begin{itemize}
\vspace*{-2mm}
\item \para{Object Remix}.
$\mathbf{Z}_{T}$ and $\mathbf{Z}_{S}$ jointly contribute to reconstructing a 3D object, so we can remix them between objects to manipulate the shape (topology) while preserving its original topology (shape); see Figure~\ref{fig:manipulation_demo}(a) for two re-synthesized coffee tables.
Figure~\ref{fig:remix} shows more results produced by remixing different tables as sources of $\mathbf{Z}_T$ (leftmost column) and $\mathbf{Z}_S$ (top row).
%
    %
%
\vspace*{-2.3mm}
\item \para{Object Interpolation.} Also, we can produce a disentangled interpolation between objects, either on $\mathbf{Z}_{T}$ or $\mathbf{Z}_{S}$, as shown in Figure~\ref{fig:manipulation_demo}(b).
From left to right, the chair (top) morphs towards the target, yet preserving its rectangular-like shape, whereas the car (bottom) becomes taller towards a truck with the same topology.
Figure~\ref{fig:interpolation} shows more results of our disentangled interpolation on topology (top) and shape (bottom).
%

%
\vspace*{-2.3mm}
\item \para{Latent Code Arithmetic}. With the learned smooth latent space,
we can exploit arithmetic operations in the latent space.
Figure~\ref{fig:manipulation_demo}(c) shows that
%
we can subtract the topology codes of two tables,~\ie, with and without a storage plate, and add the difference to the topology code of another table to augment it with a plate.
Figure~\ref{fig:arithmetic} shows more results on latent code arithmetics.
\end{itemize}
%
\vspace*{-1mm}
\supp{Please refer to the supplement for more results.}

\section{Result and Evaluation}
\label{sec:evaluation}

\subsection{Dataset and Metric}

\vspace*{-2.5mm}
\para{Dataset.} We employ 13 classes in ShapeNet~\cite{chang2015shapenet} for mesh reconstruction as~\cite{chen2019learning,chen2019bae,chen2020bsp} and adopt input voxels directly from~\cite{hane2017hierarchical} and input images from~\cite{choy20163d}.
For each reconstruction task (voxels/images), we train one model on all categories and use the same train-test split as previous works.
At inference, we directly obtain the mesh of the topology-aware neural template as a union of convexes and deform it to obtain the final mesh.
\supp{Please find details on training, testing, network architecture,~\etc, in the supplement.}

\para{Evaluation metric.}
To quantitatively evaluate the predicted mesh $\mathcal{M}$ relative to ground-truth mesh $\hat{\mathcal{M}}$,
we employ
%
(i) Light field distance (LFD); inspired by human vision system~\cite{chen2003visual,chen2020bsp}, LFD measures the visual similarity in rendered images of $\mathcal{M}$ and $\hat{\mathcal{M}}$ at different views;
(ii) Point-to-surface distance (P2F) measures the minimum distance from the sampled points of $\mathcal{M}$ to the surface of $\hat{\mathcal{M}}$; and
(iii) Chamfer distance (CD) measures the bidirectional shortest distance between the point samples of $\mathcal{M}$ and $\hat{\mathcal{M}}$.
Importantly, LFD measures the visual quality of object surfaces, whereas P2F and CD merely account for point-wise distances.
%
For all metrics, a lower value indicates a better performance.

\subsection{Mesh Reconstruction from 3D Voxels}
\label{subsec:voxel_recon}

\begin{table*}[t]
	\centering
	\caption{Quantitative evaluations on mesh reconstruction from 3D voxels. The units of LFD, P2F, and CD are 1.0, $10^{-2}$, and $10^{-3}$, resp.}
	\label{tab:quanComparison_ae}
	\vspace*{-1.5mm}
	\resizebox{1.0\linewidth}{!}{
		\begin{tabular}{C{1.1cm}|C{2.2cm}|C{1cm}|C{9mm}@{\hspace*{2.5mm}}C{9mm}@{\hspace*{2.5mm}}C{9mm}@{\hspace*{2.5mm}}C{9mm}@{\hspace*{2.5mm}}C{9mm}@{\hspace*{2.5mm}}C{9mm}@{\hspace*{2.5mm}}C{9mm}@{\hspace*{2.5mm}}C{9mm}@{\hspace*{2.5mm}}C{9mm}@{\hspace*{2.5mm}}C{9mm}@{\hspace*{2.5mm}}C{9mm}@{\hspace*{2.5mm}}C{9mm}@{\hspace*{2.5mm}}C{9mm}}
			\toprule[1pt]
			\multirow{2}*{Metric} & \multirow{2}*{Method} & \multicolumn{14}{c}{Category} \\ \cline{3-16}
			& & Mean & Plane & Bench & Cabinet & Car & Chair & Display & Lamp & Speaker & Riffle & Couch & Table & Phone & Vessel  \\ \hline
			\multirow{3}*{LFD($\downarrow$)}
			& IM-NET($256^3$) & 2918.9 & 4065.3 & 3452.7 & 1542.6 & \textbf{2069.7} & 2479.1 & 2606.2 & 6073.9 & 1763.0 & 5466.9 & 2110.7 & 2374.4 & 2109.1 & \textbf{4366.5}  \\
			& BSP-NET &3026.0 & 4287.0	& 3599. &\textbf{1489.7}&2101.1&2643.1&2602.8&6384.3&1769.8&5545.1&2170.1&2471.9&2187.7&4495.2 \\
			& Ours &\textbf{2835.0} & \textbf{3955.1}&\textbf{3329.9}&1509.1&2070.4&\textbf{2368.7}&\textbf{2460.2}&\textbf{5899.3}&\textbf{1707.1}&\textbf{5333.1}&\textbf{2043.5}&\textbf{2257.6}&\textbf{2078.6}&4366.9\\ \hline
			\multirow{3}*{P2F($\downarrow$) }
			& IM-NET($256^3$) & 0.820 & 0.597 & 0.739 & \textbf{0.749} & \textbf{0.584} & 0.876 & 0.821 & 1.543 & 1.045 & 0.794 & \textbf{0.768} & 0.930 & \textbf{0.564} & 0.864  \\
			& BSP-NET & 0.899 & 0.677 & 0.826 & 0.755 & 0.654 & 1.016 & 0.889 & 1.859 & 0.985 & 0.830 & 0.793 & 0.946 & 0.632 & 1.062  \\
			& Ours & \textbf{0.796} & \textbf{0.542} & \textbf{0.677} & 0.751 & 0.674 & \textbf{0.847} & \textbf{0.769} & \textbf{1.422} & \textbf{0.978} & \textbf{0.651} & 0.854 & \textbf{0.851} & 0.567 & \textbf{0.843}  \\ \hline
			\multirow{3}*{CD($\downarrow$)}
		    & IM-NET($256^3$) & 0.648 & 0.322 & 0.499 & 0.727 & 0.526 & 0.663 & 0.641 & \textbf{1.351} & 1.012 & 0.374 & 0.611 & 0.781 & 0.384 & 0.628 \\
		    & BSP-NET & 0.750 & 0.377 & 0.595 & 0.764 & 0.583 & 0.807 & 0.741 & 1.727 & 1.099 & 0.414 & 0.672 & 0.874 & 0.524 & 0.770 \\
		    & Ours & \textbf{0.573} & \textbf{0.259} & \textbf{0.434} & \textbf{0.651} & \textbf{0.460} & \textbf{0.581} & \textbf{0.479} & 1.442 & \textbf{0.849} & \textbf{0.268} & \textbf{0.576} & \textbf{0.674} & \textbf{0.313} & \textbf{0.550} \\ \hline
			\bottomrule[1pt]
	\end{tabular}}
	\vspace*{-2mm}
\end{table*}

\vspace*{-2.5mm}
\para{Gallery.} Figure~\ref{fig:gallery} showcases our learned neural templates (odd columns) paired with the reconstructed objects (even columns).
These results manifest that our DT-Net is able to produce topology-aware templates of various connectivity and genus specific to the target objects, and the final meshes cover a wide variety of global shapes and local structures.

\para{Quantitative evaluation.} Beyond achieving a controllable topology-aware generation of 3D meshes, we further evaluate the quality of our generated meshes against those produced by the state-of-the-art models, IM-Net~\cite{chen2019learning} and BSP-Net~\cite{chen2020bsp}.
%
%
Using the same train-test split as~\cite{groueix2018papier}, we directly leverage their pre-trained models provided in the original implementations.
%
BSP-Net and our DT-Net can directly extract meshes via a union operation of primitives from input voxels of resolution $64^3$. For IM-Net, we extract the final meshes via~\cite{lorensen1987marching} from a higher resolution input ($256^3$).
Table~\ref{tab:quanComparison_ae} reports the quantitative evaluation results,
showing that DT-Net has good performance on most categories and its overall performance also outperforms others for all metrics.
Particularly, benefited from our topology-aware neural template, DT-Net has a large improvement on object categories with high diversity in topology,~\eg, \emph{chair}.
%


\begin{figure}[t]
	\centering
	\includegraphics[width=0.999\linewidth]{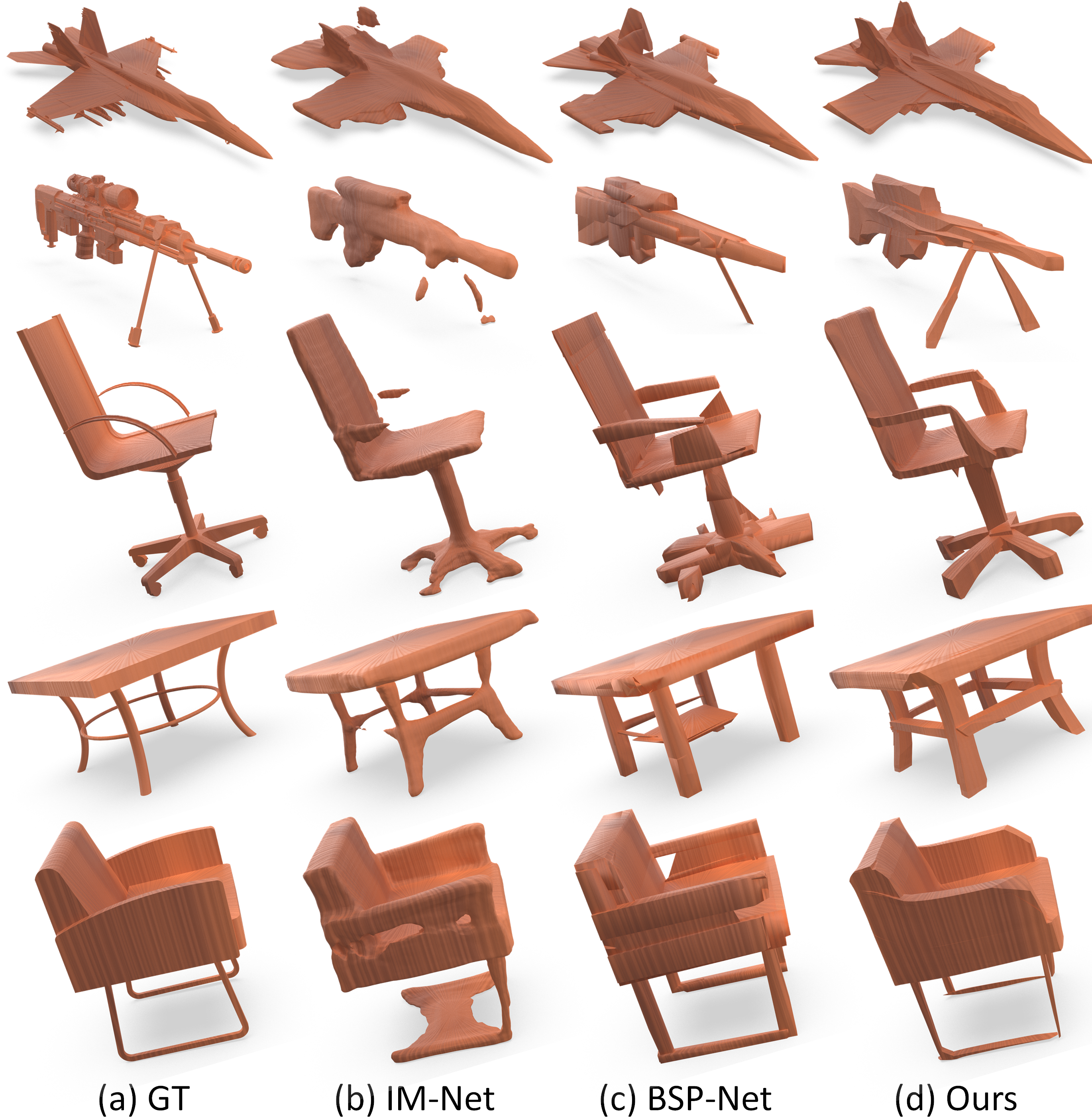}
	\vspace{-4.5mm}
	\caption{Visual comparison on mesh reconstruction from voxels.}
	\label{fig:ae_compare}
	\vspace{-4mm}
\end{figure}

\para{Qualitative evaluation.} Figure~\ref{fig:ae_compare} shows visual comparison results, revealing that
other methods tend to produce missing parts (\eg,~table's beam) and less details (\eg,~chair's pulley). In contrast, our method can produce more complete objects that are visually the closest to the targets, and our reconstructed objects exhibit more tiny local structures (\eg,~airplane and gun) and manifest various object topologies.
\supp{More results are shown in the supplement.}

\subsection{Mesh Reconstruction from Single-View Images}
\label{subsec:img_recon}

For the single-view reconstruction task,
%
we compare our method with two lines of works: (i) explicit methods: Pixel2Mesh~\cite{wang2018pixel2mesh}, AtlasNet~\cite{groueix2018papier},
and TMNet~\cite{pan19deepTopo} that directly deform a template towards the final mesh; and (ii) implicit methods: IM-Net~\cite{chen2019learning}, BSP-Net~\cite{chen2020bsp}, and DI$^2$M-Net~\cite{li2021d2im} that produce implicit surfaces.
For DI$^2$M-Net,
their authors kindly help us generate the visual results.
For other methods, we use their released implementations with the same train-test split (\ie,~80\%-20\%) and the inputs are gray-scale images.
We also noticed a very recent work~\cite{shi2021geometric}, improved from TMNet~\cite{pan19deepTopo} and we will make a proper comparison when the source code is available in the future.
%

\begin{table}[!t]
    \centering
	\caption{Quantitative results on reconstruction from 2D images. Overall, our method is better on LFD and comparable with others on distance metrics P2F and CD. Details are shown below.}
	\vspace{-1.5mm}
	\label{tab:quan_svr}
	\resizebox{1.0\linewidth}{!}{
	\begin{tabular}{C{1.5cm}|C{2.5cm}|C{1cm}|C{1cm}|C{1cm}}
		\toprule[1pt]
		\multicolumn{2}{c|}{\multirow{2}{*}{Method}}   &  \multicolumn{3}{c}{Metric}  \\ \cline{3-5}
		\multicolumn{2}{c|}{} & LFD($\downarrow$) & P2F($\downarrow$) & CD ($\downarrow$)  \\ \hline
		\multirow{3}{*}{Explicit}
        & Pixel2Mesh   & 4056.2 & 1.903 & 1.855  \\
		& AtlasNet    & 3880.9 & 1.289 & \textbf{1.041}\\
		& TMNet & \textbf{3765.5} & \textbf{1.285} & 1.149   \\ \hline
		\multirow{3}{*}{Implicit}
        & IM-NET($256^3$) & 3559.2& 1.422 & 1.497  \\
		& BSP-NET & 3426.5 & 1.354 & 1.478 \\
		& Ours & \textbf{3388.3} & \textbf{1.294} & \textbf{1.396}  \\
		\bottomrule[1pt]
	\end{tabular}}
\vspace{-1.5mm}
\end{table}

\begin{figure*}[t]
	\centering
	\includegraphics[width=0.99\linewidth]{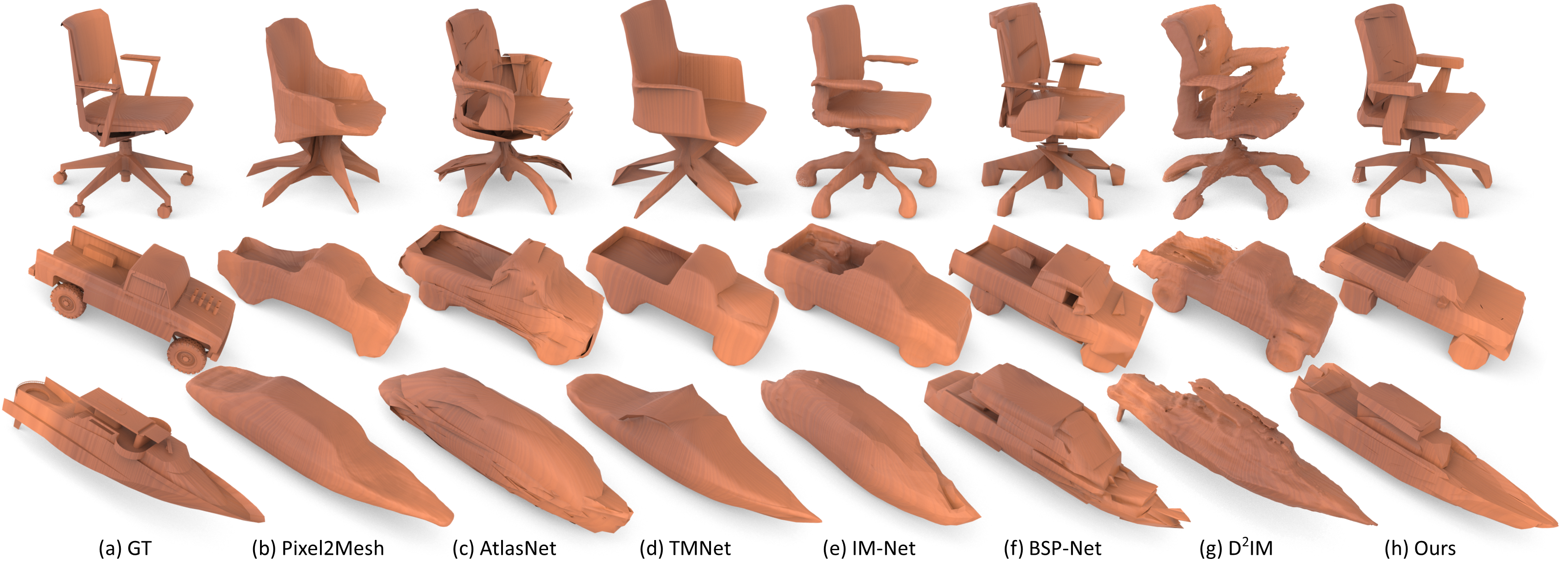}
	\vspace{-3mm}
	\caption{Visual comparison on 3D mesh reconstruction from 2D images.}
	\label{fig:svr_compare}
	\vspace{-4mm}
\end{figure*}
\para{Quantitative evaluation.}
Table~\ref{tab:quan_svr} lists the overall results, showing that our method consistently outperforms the other implicit methods in terms of LFD, P2F, and CD.
Note that we did not include DI$^2$M-Net, since it requires an additional camera pose as input for the training.
On the other hand, comparing with the explicit methods, our method is better on LFD and comparable on the distance-based metrics CD and P2F; this might be attributed to their CD-wise regularization in the training. Also, distance-based metrics may not be ideal for measuring the quality of the reconstructed meshes (see~\cite{chen2020bsp,jin2020dr}), 
as evidenced by the visual comparison results in Figure~\ref{fig:svr_compare}.
\supp{We also show the detailed results on individual categories in the supplement.}

\para{Qualitative evaluation.} Figure~\ref{fig:svr_compare} shows the visual comparison results.
Referring to the ground-truth meshes (a), explicit methods (b-d) are typically hard to adapt objects of various genus, therefore further confirming our motivation for topology-aware template formulation.
On the other hand, implicit methods (e-g) can describe the topology flexibly, yet tend to produce over-smooth or noisy surfaces with less details,~\eg, chair's armrest and boat's hull. In contrast, our method (h) can produce high-quality meshes, in which
the surfaces exhibit smooth and sharp features simultaneously.
\supp{More visual results are in the supplement.}

\subsection{Model Analysis and Discussion}
\label{subsec:ablation}
\begin{figure}[t]
	\centering
	\includegraphics[width=0.99\linewidth]{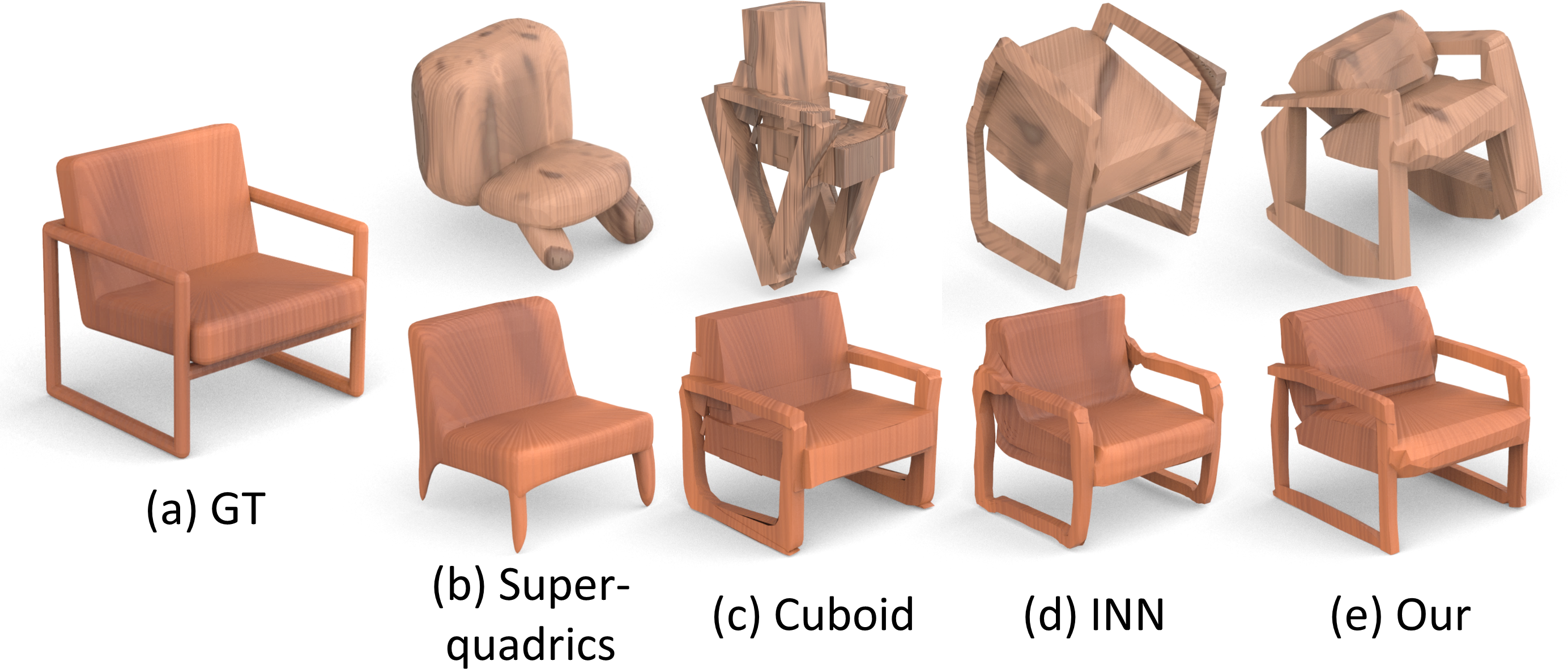}
	\vspace{-2mm}
	\caption{Given the reference mesh (a), we may use alternative representations (b-c) to compose the neural template or use INN (d) for shape deformation. Our method (e) shows better results.}
	\label{fig:analysis}
	\vspace{-3mm}
\end{figure}

\vspace{-2mm}
\para{Framework analysis.} \
We first verify the framework design of DT-Net.
%
%
%
In Figure~\ref{fig:analysis},
%
%
we compare the usage of other primitives~\cite{tulsiani2017learning,paschalidou2019superquadrics} such as superquadrics (b) and cuboids (c), vs. our convexes (e) for composing the topology-aware neural templates.
Figure~\ref{fig:analysis} (d) shows results when using another invertible neural network (INN)~\cite{paschalidou2021neural} to implement the shape deformation module; \supp{see details and evaluation results in supplement.}
%
%
Generally,
we take a generic design for the topology formation and shape deformation modules (see Section~\ref{subsec:framework}), meaning that
we may use alternative implementations, yet our current choices provide better topological approximations and lead to better reconstructions.

%

\vspace{-0.5mm}

\para{Visualization of the topology space.} \
To show the smoothness and meaningfulness of the learned topology space,
we produce a visualization of the TSNE embedding
for $\mathbf{Z}_{T}$ on chairs.
From the visualization shown in \supp{Figure~{8} of the supplement}, we can see that
DT-Net can learn a smooth embedding space for objects of varying topological structures and objects of similar topologies are closely clustered.
\vspace{-0.5mm}

\begin{figure}[t]
	\centering
	\includegraphics[width=0.93\linewidth]{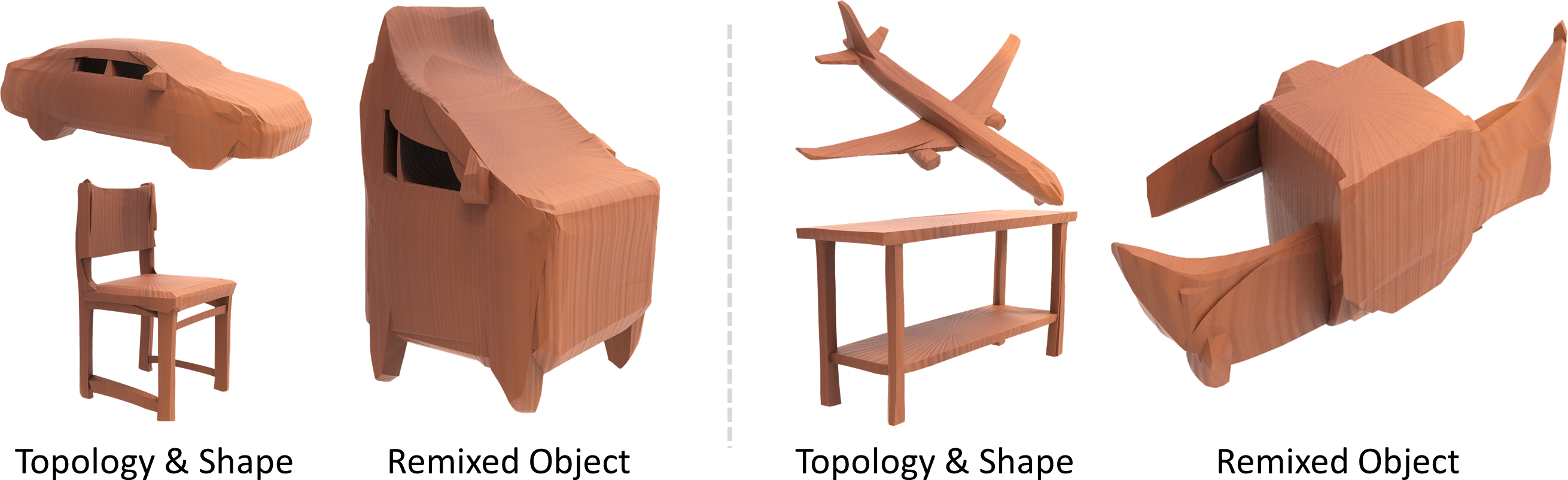}
\vspace{-2mm}
	\caption{Cross-category object remix between different classes.}
	\label{fig:cross-mix}
	\vspace{-4mm}
\end{figure}

\para{Cross-category manipulation.} \
Since our model is trained on multiple object categories, we may conduct object remixing across different categories; see the results in Figure~\ref{fig:cross-mix}. Interestingly, we can obtain a chair-like car, which follows the car's topology and the chair's shape.
\supp{We show more cross-category results in the supplement.}
\vspace{-0.5mm}

\para{Limitation and Discussion.} \
First, like most of the previous methods on 3D mesh generation,
it is still very challenging to produce objects of extremely complex and fine structures; \supp{see the supplement}. In the future, we aim to further formulate the topology-aware neural template in a hierarchical manner and deform it in a part-wise manner for more fine-grained reconstructions and controls.
%
%
Also, since DT-Net is built on a reconstruction task, the generated new objects are still limited to the diversity of the given objects. We would like to extend it into an unsupervised generation framework and take into account voice, text, or other input modality for more intuitive object manipulations.
\section{Conclusion}
\label{sec:conclusion}

We presented a novel framework called DT-Net that enables a topology-aware mesh reconstruction and promotes mesh generation with disentangled controls.
A key design is to learn to form a topology-aware neural template specific to each input then deform it to reconstruct a detailed 3D object.
This scheme decouples the 3D reconstruction process into two sub-tasks, effectively accommodating for the variations in topology.
Importantly, our new design provides a disentangled representation of topology and shape in the latent space, enabling controllable object generations by manipulating the learned topology code and shape code, which
are not achievable by the existing reconstruction methods.
Extensive experiments also manifest that our method produces high-quality meshes with
diverse topologies and fine details, performing favorably over the state of the arts.
%


\para{Acknowledgments.} \
\new{We thank anonymous reviewers for the valuable comments.
This work is supported by the Research Grants Council of the Hong Kong Special Administrative Region (Project No. CUHK 14206320 \& 14201921).}

{\small
\bibliographystyle{ieee_fullname}
\bibliography{egbib}
}

\end{document}